# Knowledge Map: Toward a new approach supporting the knowledge management in Distributed Data Mining

Nhien-An Le-Khac
School of Computer Science
and Informatics
University College Dublin
Dublin 4, Ireland
Email: an.lekhac@ucd.ie

Lamine Aouad
School of Computer Science
and Informatics
University College Dublin
Dublin 4, Ireland
Email: lamine.aouad@ucd.ie

M-Tahar Kechadi
School of Computer Science
and Informatics
University College Dublin
Dublin 4, Ireland
Email: tahar.kechadi@ucd.ie

*Abstract*—Distributed data mining (*DDM*) deals with the problem of finding patterns or models, so-called knowledge, in an environment with distributed data and computations. Today, a massive amounts of data which are often geographically distributed and owned by different organisation are being mined. As consequence, a large mounts of knowledge are being produced. This causes problems of not only knowledge management but also visualization in data mining. Besides, the main aim of *DDM* is to exploit fully the benefit of distributed data analysis while minimising the communication. Existing *DDM* techniques perform partial analysis of local data at individual sites and then generate a global model by aggregating these local results. These two steps are not independent since naive approaches to local analysis may produce an incorrect and ambiguous global data model. The integrating and cooperating of these two steps need an effective knowledge management, concretely an efficient map of knowledge in order to take the advantage of knowledge mined to guide mining the data.

In this paper, we present "knowledge map", an representation of knowledge about knowledge mined. This new approach aims to manage efficiently knowledges mined in large scale distributed platform such as Grid. This knowledge map is used to facilitate not only the visualization, evaluation of mining results but also the coordinating of local mining process and existing knowledge to increase the accuracy of final model.

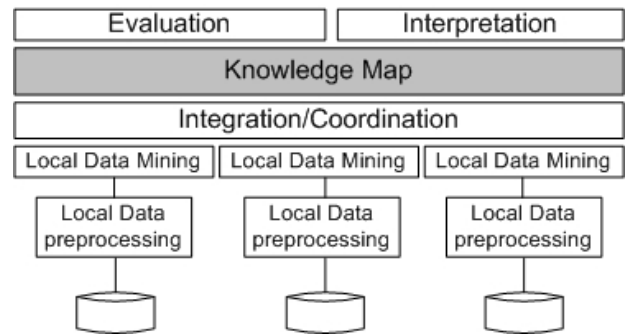

Fig. 1. ADMIRE organization

## I. INTRODUCTION

Today a deluge of data are collected from not only science fields but also industry and commerce fields. In this context, distributed data mining (*DDM*) techniques have become necessary for analyzing huge and multi-dimensional datasets distributed in large number of sites. Massive amounts of data that are being gathered and mined and a large amounts of the mining results is as consequence produced. This phenomenon conducts to the problem of managing these results so-called knowledge mined. When the knowledge mined is exploded, the presentation of them gets more complex. Thus, knowledge mined management affects firstly to the visualization of mining results. A visual data mining system[9] must be syntactically simple to be useful: simple to learn, to apply, to retrieve and to execute. So, an useful visual data mining system should be based on an efficient knowledge management. This is even more critical when knowledge located on different sites owned by different organisations. Besides, existing *DDM* techniques is based on performing partial analysis on local data at individual sites and then generating a global model by aggregating these local results. These two steps are not independent since naive approaches to local analysis may produce an incorrect and ambiguous global data model. In order to take the advantage of knowledge mined to guide mining the data, *DDM* should have a coordinating and integrating view of knowledge mined that needs a efficient map of these knowledge. Briefly, an efficient management of knowledge mined is one of important key factors affecting the successfulness of these tasks.

We present, in this paper, a "knowledge map" whose principal role is to manage knowledge mined in large scale distributed system. By proposing this knowledge map, we aim to provide a simple and efficient way to handle the large amount of knowledge mined collected from Grid environments. This map helps to retrieve quickly knowledge needed with a high accuracy and to coordinate them in order to guide mining the data. We aim also offer an effective knowledge map to facilitate the visualization for data mining. This knowledge map is a part of ADMIRE[6] (Fig.1), a framework based on Grid platform for developing novel *DDM* techniques to deal with very large and distributed heterogeneous datasets in both commercial and academic applications. This framework is being developed in the Department of Computer Science at University College Dublin. The rest of this paper is organized as follow: Section 2 deals with background of knowledge representation and knowledge map concept as well as related project then we will present the architecture of our knowledge map in section 3. Section 4 presents knowledge map's operations and our evaluations of this approach will be held in section 5. Finally, we conclude on Section 6.

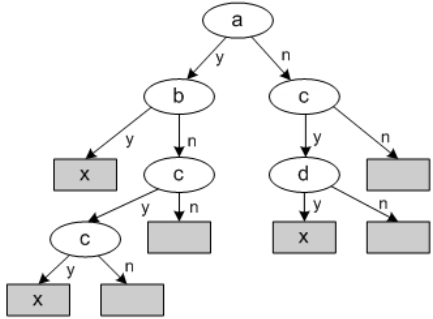

Fig. 2. Knowledge representations

## II. BACKGROUND

In this section, we present firstly some methods for representing knowledge in data mining. Next, we will discuss the concept of knowledge map and the using of knowledge map in managing the knowledge as well as some research about it.

### A. Knowledge representation

There are many different ways for representing mined knowledge in literature that could be listed as decision tables, decision trees, classification rules, association rules, instance-based and clusters.

Decision table is one of the simplest ways of representing knowledge. Its columns contains set of attributes including the decision one and its rows is knowledge elements (Fig.2a). This structure is simple but it might waste storage space because of involving unused attributes. Decision tree approach (Fig.2b) is based on "divide-and-conquer" concept where each node of it involves testing a particular attribute and leaf nodes give a classification that applies to all instances that reach the leaf. This approach has to however deal with missing value problem. Classification rule[3] is a popular alternative to decision tree. This approach uses production rule[],so-called cause-effect relationship, to express knowledge. A decision tree is used to represent the relationship between rules. Association rules[3] is a kind of classification rule except that they can predict any attribute and this gives them the freedom to predict combinations of attributes too. Moreover, association rules are not intended to be used together as a set, as classification rules are. The instance-based knowledge representation uses the instances themselves to represent what is mined rather than inferring a rule set and store it instead. The problem is that they do not make explicit the structures that they are mined. In the cluster approach, mined knowledge takes the form of a diagram that shows how the instances fall into clusters. There are many kinds of cluster representation such as space partitioning, Venn diagram, table, tree, etc. Clustering[3] is often followed by a stage in which a decision tree or rule set is inferred that allocates each instance to the cluster in which it belongs. An remarkable notice of knowledge representation method is that production rule is used in most cases.

### B. Knowledge map concept

A knowledge map is generally a visual representation of "knowledge about knowledge" rather than of knowledge itself[2][4][10]. It basically helps to detect the sources of knowledge and structure of knowledge by representing the elements and structural links of application domains. Some kind of knowledge map structure can be found in literature are: hierarchical/radial knowledge map, networked knowledge map, knowledge source map and knowledge flow map.

Hierarchical knowledge map, so-called concept maps[8] provide one model for the hierarchical organization of knowledge: top-level concepts are abstract with few characteristics. Concepts on the level below have detailed individual traits of the super concept. The propositions between concepts can represent any type of relations as "is part of", "influences", "can determine", etc. A similar approach is radial knowledge map or mind map[1] which consists of concepts that are linked through propositions. However, it has radially organized. Networked knowledge map is also called causal map which is defined as a technique "for linking strategic thinking and acting, helping make sense of complex problems, and communicating to oneself and others what might be done about them"[1]. This approach is normally served for systematizing knowledge about cause and effects. Knowledge source map[4] is a kind of organizational charts that does not describe functions,responsibility and hierarchy, but expertise. It helps to detect experts in a specific knowledge domain. The last one, knowledge flow map[4] represents the order in which knowledge resources are and should be used rather than a map of knowledge.

Few research on knowledge map that can be cited as [5][7]. However, these research were not in the context of *DDM*. Until now, at our best knowledge, in spite of a deluge of knowledge mined from *DDM* applications, there is no work on knowledge map that manages and coordinates these knowledge to facilitate the visualization of mining results as well as to support *DDM* tasks yet. This is also the motivation of our research.

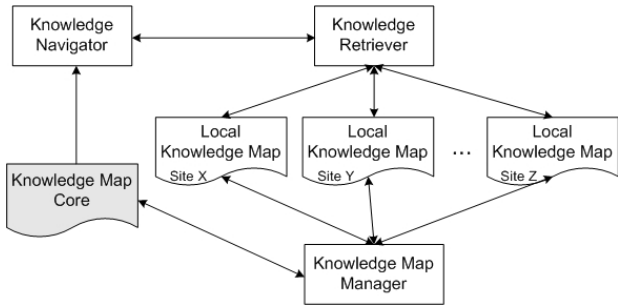

Fig. 3. Knowledge map system

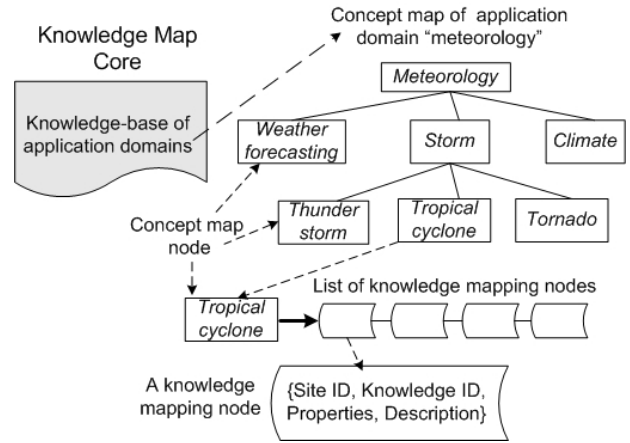

Fig. 4. Knowledge map core structure

## III. ARCHITECTURE OF KNOWLEDGE MAP

An knowledge map does not attempt to systematize the knowledge itself but rather to codify "knowledge about knowledge". In our context of facilitating *DDM* by supporting users in coordinating and cooperating knowledge mined, the elementary objective of knowledge map architecture is to offer an efficient way to retrieve the these knowledge. In this section, we propose the architecture of the knowledge map system as shown in Fig.3,4,5 to achieve this purpose. A well-organized knowledge map structure will help searching process retrieve rapidly the knowledge they needed. In our approach, knowledge map is consisted of main components: knowledge navigator, knowledge map core, knowledge retriever, local knowledge map and knowledge map manager (Fig.3).

### A. Knowledge navigator

Normally, users might not exactly know the knowledge they are looking for. Thus, knowledge navigator component is firstly responsible for guiding users to traverse the knowledge map in order to detail application domains related as well as to determine knowledge mined needed. The result of this task is not contents of knowledge but its related information such as data mining task used, data type of dataset mined, a brief description of this knowledge and its location. For example, user want to retrieve knowledge mined about tropical cyclone. Application domain "meteorology" is used by this component to navigate user to tropical cyclone area and then a list of related knowledge with their information will be supplied. Next, basing on this knowledge information, users will decide which knowledge and its location as well to be retrieved. Retrieving is the second role of this component. It will interact with retriever component to collect all of knowledge related in detail from locations chosen. In distributed system, knowledge navigator component is implanted in each local site.

### B. Knowledge map core

This component (Fig.4) includes a knowledge-base storing a set of application domains. Each application domain has a hierarchical structure, so-called concept map[8], with a limited number of level. Each node in this structure contains a list of knowledge mapping nodes which represent for a knowledge mined from local sites in the distributed system. A knowledge mapping node is composed by four elements: identify of site (Site ID) where this knowledge is stored, identify of this knowledge (Knowledge ID) in this site, properties and description of this knowledge. Knowledge properties include important features of this knowledge such as data type, dimension, data mining task executed, data size, quality, etc. and knowledge description is equivalent with a brief report about this knowledge.

As shown in Fig.4 as an example, knowledge-base contains an application domain named "meteorology" which includes sub-domains such as "weather forecasting", "storm" and "climate". And then, "thunder storm", "tropical cyclone" and "tornado" are parts of "storm". Knowledge mapping list of "tropical cyclone" node manages information about knowledge mined from hurricane data, typhoon data, etc. For example, the first node of this list contains information as {(**SiteID**=*"pcrgcluster.ucd.ie"*), (**KnowledgeID**=*"16"*), (**Properties**=*"data type: numeric-interval*, dimension: *12*, data mining task: *association rules*, data size: *60Gb"*), (**Description**=*"knowledge mined from Hurricane Isabel data []"*)}. Basing on these information, users could determine which knowledge mined they want to retrieve.

By proposing knowledge map core, we aim not only to help to detect the sources of knowledge and information but also to represent them in the relationship among concepts of application domain. This component could be implanted in a master site assigned/voted from set of local sites used or in all local sites depending on topology of distributed system. Knowledge map core also offers capacity of determining one or a set of knowledges belonging to many sub-domains of an application domain. The creation and maintenance of this component as well as its operations will be presented in the section 4.

### C. Knowledge retriever

The role of knowledge retriever is to go seeking all knowledge mined needed. The realization of this task bases on information provided by users after navigating on application domains. This component is looked like a search engine which interacts with related local sites to retrieve information and return knowledge acquired in detail.

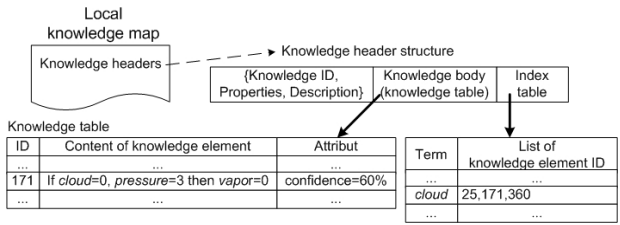

Fig. 5. Organization of local knowledge map

### D. Local knowledge map

Local knowledge map is resided at each local site where knowledge was built after mining process. Without losing the generality, we can assume that a knowledge is composed by a set of elements, so-called knowledge elements. This component not only provides an efficient way to manage these knowledge but also represents the relationship between elements of each knowledge. Local knowledge map is a set of knowledge headers. Each header is consisted of three parts: management attribute, pointer to knowledge table and pointer to index table (Fig.5). Management attribute contains important information about this knowledge: identify (Knowledge ID), properties and description of this knowledge. These information were explained in the section *Knowledge map core* above and they are also information stored in related knowledge mapping node in the knowledge map core.

Knowledge table contains knowledge mined itself. Each entry of this table represents for an element of this knowledge and is composed by three factors: identify of element (EID), content and attribute of element. For example, suppose that knowledge mined of an association rule task is represented by the production rule[], so the content of an element of this knowledge is an "IF-THEN-" phrase and its attributes are support and confidential[3].

Index table is a data structure that maps terms to the knowledge elements that contain them. For example, the index of a book maps a set of selected terms to page numbers. There are many different types of index described in literature. In our approach, index table is based on *inverted file*[?] technique because it is one of the most efficient index structure [?]. This index table consists of two parts: terms and a collection of lists, one per term, recording the identifiers of the knowledge elements containing that term. For example (Fig.5), we assume that the term *"cloud"* exists in rules *25, 171, 360* so its list is {*25, 171, 360*}. This index table also express the relationship between terms and knowledge elements containing them. By using this table, knowledge elements related to searching terms will be retrieved by the intersection of their list, e.g. list of term *"pressure"* is *20, 171* so the ID of knowledge element that contains *"cloud"* and *"pressure"* is *171*.

### E. Knowledge map manager

This component, as its name, is responsible for managing and coordinating the local knowledge map and the knowledge map core. At local knowledge map side, knowledge map manager allows to create a set of knowledge headers and index tables bases on existing knowledge elements. It also supports to build the knowledge table if it is not exist yet. It manages local knowledge map operations on the header, the index table and the knowledge table such as adding, editing and deleting. Knowledge map manager also manages the creation and maintenance of knowledge map core. An important role of this component is to keep the coherence between local knowledge map and knowledge map core.

## IV. KNOWLEDGE MAP OPERATIONS

In this section, we present three basic operations on knowledge map system: creation the local knowledge map and knowledge map core, retrieving the knowledge and maintenance of knowledge map system.

### A. Creation

*Local knowledge map*: Firstly, the knowledge elements of the same knowledge mined are restructured in its knowledge table. Next, index table for each knowledge is constituted by scanning its knowledge table and its knowledge header will be then built up basing on these tables. Information of these headers will be sent to knowledge map core to create or to update the related knowledge mapping.

*Knowledge map core*: Some popular application domains are firstly predefined and their hierarchical nodes are built. Note that system also allows user to add new application domains at runtime. At each hierarchical node, knowledge mapping list then will be created and adds knowledge mapping node with information basing on knowledge headers from local knowledge maps.

### B. Knowledge retrieval

This is one of the most important operations. The searching process begins with navigating users to browse on application domains and then they will determine one or a set of hierarchical nodes (concept map nodes) of an application domains that they need to seek knowledge. If they only chose one application domain node, information of knowledge in its knowledge mapping list will be resumed. Otherwise, if more than one hierarchical node chosen, an intersection of their knowledge mapping lists is executed to resume information of knowledges related to all sub-domains from hierarchical nodes. Then, users choose some knowledge basing on their information (location, properties and description) to retrieve their contents in detail. Moreover, users are also asked for entering keywords of related terms contained in knowledge. knowledge map system uses their (*SiteID, KnowledgeID*) and these keywords to seek knowledge elements on related local sites. At each local site, headers of related knowledge is examined and its index table is used to determine knowledge elements related to supplied keywords. Next, set of satisfied knowledge elements will be collected and returned to requiring site.

For example, users firstly want to seek knowledge about hurricane, so the application domain "meteorology" is used to navigate user to tropical cyclone area and then a list of knowledge about hurricane, typhoon, etc. with their information will be resumed. In this case, only one application domain node "tropical cyclone" is chosen. Next, basing on this knowledge information, users decide to retrieve knowledge of association rule task on hurricane Isabel data [] located on site X and site Y. They also enter keywords of concerned terms as *"pressure"* and *"cloud"*. Basing on these information, knowledge map system will searching on local knowledge maps of site X and of site Y. Index tables of chosen knowledge on each site are exploited to retrieve knowledge elements related to required terms. We assume that site X contains local knowledge map as in Fig.5 and lists of *"pressure", "cloud"* are as in the example of section *III.D*, so knowledge element with *ID = 171* will be taken. The same action is executed at site Y and then a set of related knowledge elements will be collected and returned.

## C. Maintenance

In *DDM* environment, knowledge mined is contributed continuously. Therefore, this function is designed to keep the knowledge map system up to date: how to index the new incoming knowledge in the existing knowledge map, how to update related information of existing knowledge, how to assure the coherence of knowledge information between local knowledge map and knowledge map core.

*Adding a new knowledge*: the knowledge table and index table of this new knowledge are built up and then its header with be added in the knowledge header set at this local site. In the next step, its application domain is chosen basing on the knowledge-base of application domain in knowledge map core. If its hierarchical node (concept map node) has existed, a new knowledge mapping node will be produced and be added in the knowledge mapping list at the related concept map node. If not, a new application domain node will be created and then its knowledge mapping list will be set up with a knowledge mapping node of this new knowledge as the first element. If an appropriate application domain does not exist, knowledge map system allows to define a new application domain and its hierarchical sub-domain.

*Updating of an existing knowledge*: an existing knowledge might be updated its application domain or its hierarchical sub-domain. In this case, its knowledge mapping node is removed from its list and is added to the list of new sub-domain.

*Assuring the coherence*: At knowledge map core, only updating the application domains as above is allowed. The updating of information of a knowledge map node is prohibited. These information is only updated from local knowledge map. This operation is moreover atomic.

## V. EVALUATION

To the best of our knowledge, there exist no knowledge map which helps to manage mined knowledge in *DDM*. Thus, it is difficult to evaluate our approach by comparison with another one. Therefore, we estimate this new approach by evaluate different aspect of their architecture in the context of supporting the management, mapping, representation and retrieval knowledge as evaluation criteria.

First of all, we evaluate the complexity of this knowledge map system. Two important operations estimated are searching/retrieving knowledge and maintenance the system. The first operation includes two parts: searching at knowledge map core and searching/retrieving knowledge in local knowledge map at each local site. Let N be the number of application domains at n is the number of application sub-domains of each application domain, so the complexity of the first part is *O(logN + logn)* because application domains are normally indexed followed B_tree or B+_tree.

In the same way, let M be the number of knowledge headers and m be the number of required terms of related knowledge, the complexity of searching/retrieving knowledge at local knowledge map is *O(logM + m x CIT)* where CIT is the searching cost of each term in the related index table. This cost was analyzed as in []. Next, we estimate the complexity of maintenance the system via the adding cost of a new knowledge, other operations as updating,etc. have the same evaluation. Using the same symbol as in the first operation, the complexity of adding a new knowledge is *O(m x CIT + logN + logn + CL)* where CL is the complexity of adding an element in knowledge mapping list of an application domain node.

Next, we estimate the knowledge map architecture. Firstly, the structure of application domains bases on architecture of concept map[] so it can benefit advantages of this model []. We can also avoid the problem of semantic ambiguous as well as limit the searching domain to improve the seeking speed with higher accuracy. Secondly, the division of knowledge map in two main components (local and core) has some advantages:

- core component acts as a map of knowledge and it is moreover a representation of knowledge about knowledge when combining with local knowledge headers;
- avoiding the movement of whole stored knowledge to one master site (or server) it is impossible in large scale distributed system as Grid.

By using the index table, local knowledge map is moreover a map of knowledge elements by representing relationships between terms and knowledge elements containing them. Finally, our new approach offers a knowledge map with flexible and dynamic architecture:

- users could update easily application domains and their sub-domains;
- knowledge mapping element of the same knowledge is allowed to exist at different application domain nodes because it could be difficult to force the users in assigning knowledge to only one existing sub-domain;
- knowledge table at each local knowledge map allow to store knowledge element not only in production rule but also in other forms. In that case, its contain could be a pointer to a text, a graphic, etc.

## VI. CONCLUSION

In this paper, we present the architecture of a knowledge map. This innovation structure aims to manage effectively knowledge mined in large scale distributed platform. The purpose of this research is to provide a knowledge map to facilitate the visualization of mining results as well as to support the *DDM* tasks. Throughout estimations of each component and it function, we can conclude that knowledge map system has an efficient and flexible architecture. It satisfies the need of managing, retrieving mined knowledge of *DDM* in large distributed environment.

This knowledge map is being improved and will be integrated in ADMIRE framework. Experiences of whole system will be executed to evaluate it with real, large scale applications.